\documentclass{article}
\usepackage{amssymb}
\usepackage{kotex} 
\usepackage{graphicx}
\usepackage{subfigure}
\usepackage{adjustbox}
\usepackage{siunitx}
\usepackage{booktabs}

\usepackage[preprint]{corl_2025} 

\title{Scoop-and-Toss: Dynamic Object Collection for Quadrupedal Systems}

\author{
  Minji Kang \\
  Department of Computer Science\\
  Hanyang University, South Korea \\
  \texttt{alswl7763@hanyang.ac.kr} \\
  \And
  Chanwoo Baek \\
  Department of Computer Science\ \\
  Hanyang University, South Korea \\
  \texttt{bcw0430@hanyang.ac.kr} \\
  \And
  Yoonsang Lee \\
  Department of Computer Science\\
  Hanyang University, South Korea \\
  \texttt{yoonsanglee@hanyang.ac.kr} \\
}

\begin{document}
\maketitle


\begin{abstract}
    
Quadruped robots have made significant advances in locomotion, extending their capabilities from controlled environments to real-world applications.
Beyond movement, recent work has explored loco-manipulation using the legs to perform tasks such as pressing buttons or opening doors.
While these efforts demonstrate the feasibility of leg-based manipulation, most have focused on relatively static tasks.
In this work, we propose a framework that enables quadruped robots to collect objects without additional actuators by leveraging the agility of their legs.
By attaching a simple scoop-like add-on to one leg, the robot can scoop objects and toss them into a collection tray mounted on its back.
Our method employs a hierarchical policy structure comprising two expert policies—one for scooping and tossing, and one for approaching object positions—and a meta-policy that dynamically switches between them.
The expert policies are trained separately, followed by meta-policy training for coordinated multi-object collection.
This approach demonstrates how quadruped legs can be effectively utilized for dynamic object manipulation, expanding their role beyond locomotion.

\end{abstract}




\section{Introduction}
	

Recent years have seen a rapid expansion in the applications of quadruped robots.
Research has focused initially on their locomotion capabilities,  with many studies training quadruped robots using deep reinforcement learning (DRL) in controlled environments \cite{hwangbo_learning_2019,DBLP:conf/rss/HaarnojaHZTTL19}.
These studies have expanded to include locomotion in the wild across challenging terrain \cite{lee_learning_2020,DBLP:conf/rss/KumarFPM21,miki_learning_2022}, 
navigation in complex 3D spaces with obstacles \cite{DBLP:conf/iros/ChengLPLY24,xu_dexterous_2024}, and ensuring safe locomotion in risky environments \cite{schneider_learning_2024,DBLP:conf/iros/XiaoZZZ24}.

While early research primarily focused on locomotion, manipulation capabilities have since emerged as an important complementary area.
One approach has been to integrate a robotic arm onto the quadruped platform, allowing the robot to maintain mobility while executing diverse manipulation behaviors~\cite{ma_combining_2022, fu_deep_2023}.  
However, mounting an arm increases system complexity, weight, and energy consumption, leading to significant trade-offs.

As an alternative direction, recent studies have explored techniques that leverage the robot's legs during locomotion to perform manipulation tasks.  
Some approaches use a single leg to kick a ball, open doors, or press buttons~\cite{ji_hierarchical_2022, cheng_legs_2023}, while others propose attaching grippers to the legs to grasp and manipulate objects~\cite{tsvetkov_novel_2022, lin_locoman_2024}.  
Many of these studies have focused on relatively static tasks, where both the legs and manipulated objects move slowly, although some have explored more dynamic scenarios such as ball kicking.  
While these efforts represent important progress, there remains substantial untapped potential to exploit the agility and force-generation capabilities of quadruped legs, which are sufficiently fast and powerful to support rapid locomotion.

\begin{figure}
  \centering
  \subfigure[]{
    \includegraphics[trim=0 0 230 0, clip, width=.46\linewidth]{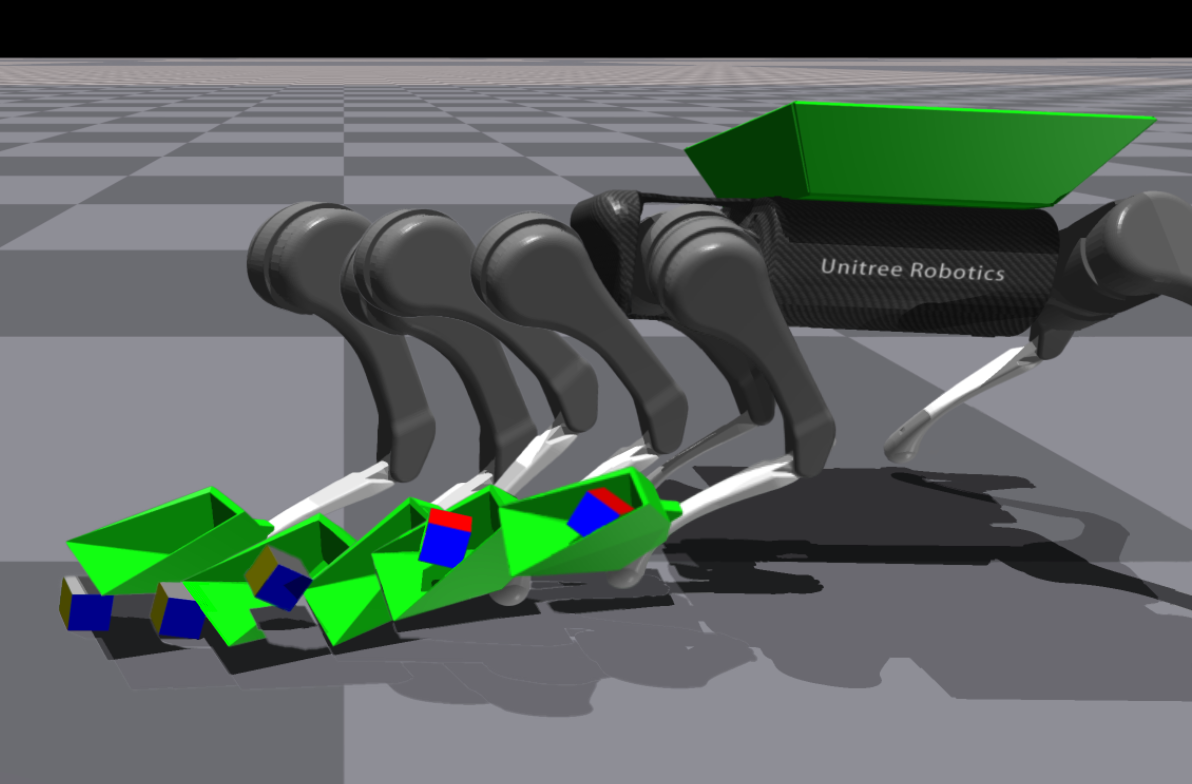}
    }
  \subfigure[]{
    \includegraphics[trim=0 30 0 0, clip, width=.46\linewidth]{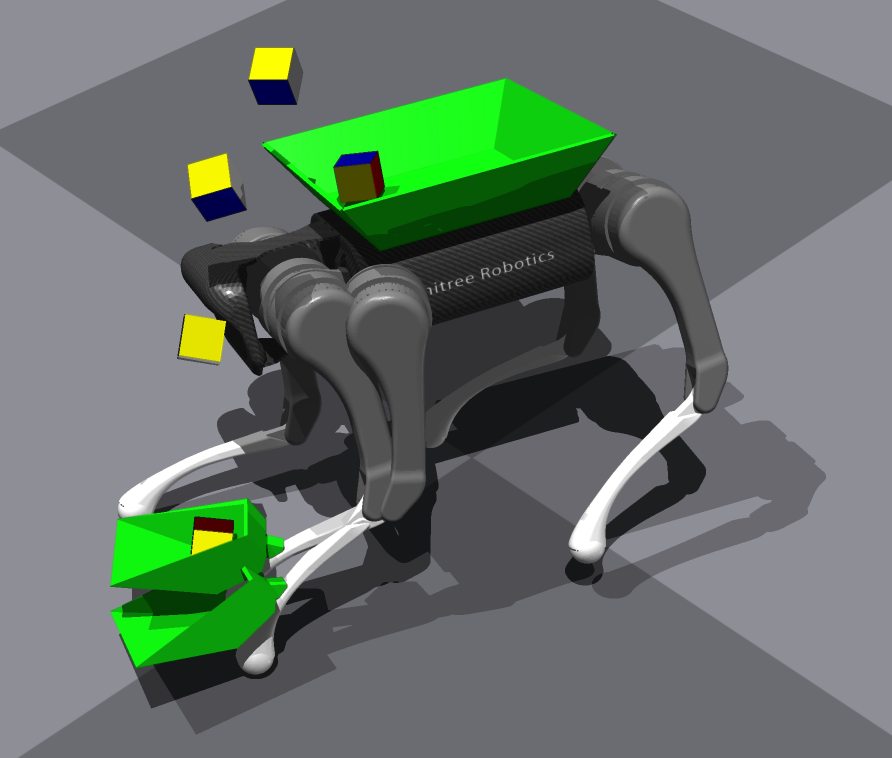}
    }
  \caption{Our proposed Scoop-and-Toss framework enables the robot to perform object collection using simple add-ons (green): a scoop for picking up objects and a collection tray for storing them.
The robot strikes objects on the ground with the scoop, rolling them inside (a), and then tosses them upward into the tray mounted on its back (b).}
\label{fig:teaser}
\end{figure}


Building on the inherent agility and strength of quadruped legs, we propose a framework that enables quadruped robots to collect objects without requiring additional actuators.
As shown in Figure~\ref{fig:teaser}, our approach involves attaching a simple add-on, such as a scoop-like extension, to one of the robot's legs, allowing it to ``scoop" objects and ``toss" them into a collection tray mounted on its back.
This design leverages the robot’s locomotion capabilities to collect and load objects, expanding the role of quadruped legs beyond locomotion and relatively static manipulation.



Our method employs a hierarchical policy structure comprising expert policies and a meta-policy, trained in two stages.
In the first stage, expert policies are trained separately: the scoop-and-toss policy enables the robot to scoop up an object within range and toss it into the tray, while the approach policy allows it to navigate toward a given object.
Once the experts are trained, the second stage trains the meta-policy, which dynamically selects between them based on the robot and object states.
This hierarchical framework allows the robot to alternate seamlessly between locomotion and manipulation behaviors, enabling efficient collection of multiple objects scattered throughout the environment.
By leveraging simple mechanical extensions and learned policies, our approach shows how quadrupeds can perform dynamic loco-manipulation without additional actuators.

\section{Related Work}

\paragraph{Learning Locomotion and Locomotion-Based Skills}

Research in quadrupedal locomotion has achieved robust terrain traversal~\cite{hwangbo_learning_2019,DBLP:conf/rss/HaarnojaHZTTL19,lee_learning_2020,DBLP:conf/rss/KumarFPM21,miki_learning_2022} and perception-aware navigation in complex 3D environments~\cite{DBLP:conf/iros/ChengLPLY24,xu_dexterous_2024}.
Beyond standard locomotion, recent work has tackled dynamic challenges such as parkour~\cite{zhuang2023robot,cheng2023parkour} and bipedal behaviors on quadrupedal hardware~\cite{li_learning_2024,su_leveraging_2024}, expanding their versatility.
Other efforts address risk-aware control~\cite{schneider_learning_2024}, perturbation robustness~\cite{DBLP:conf/iros/XiaoZZZ24}, and animal motion imitation for improved naturalness and efficiency~\cite{DBLP:conf/rss/PengCZLTL20}, laying the groundwork for more dynamic and interactive locomotion systems.

\paragraph{Loco-Manipulation with Quadrupeds}

A wide range of methods aim to extend quadrupedal robots' capabilities beyond locomotion by integrating manipulation. One class of approaches mounted robotic arms onto mobile bases to enable manipulation while maintaining mobility~\cite{bellicoso_alma_2019,zimmermann_go_2021,ma_combining_2022}.
These systems explored both model-based planning methods~\cite{mittal_articulated_2022,sleiman_versatile_2023,ferrolho_roloma_2023,rigo_hierarchical_2024} and learning-based control frameworks~\cite{fu_deep_2023,portela_learning_2024}, aiming to coordinate locomotion and manipulation in a unified system. Although these methods have expanded the range of possible tasks, the addition of robotic arms often introduced mechanical complexity, increased payload, and extensive modeling challenges.

To address these challenges, recent research has increasingly explored utilizing the legs themselves for manipulation tasks, such as kicking, dribbling, blocking a ball, pressing buttons, or pushing doors and other objects~\cite{ji_hierarchical_2022,ji_dribblebot_2023,huang_creating_2023,cheng_legs_2023,he_learning_2024,huang_hilma-res_2024}.
Leg-mounted manipulators have been introduced to enhance manipulation capabilities while preserving locomotion ability~\cite{tsvetkov_novel_2022,arm_pedipulate_2024,lin_locoman_2024}. These studies collectively demonstrate a growing trend toward directly integrating mobility and manipulation within legged systems.

Our work builds on this line by proposing a dynamic object collection framework based on simple passive extensions and a hierarchical policy structure, enabling efficient manipulation without requiring additional actuators or complex model-based integration. While prior work has demonstrated object loading using leg-mounted gripper-like devices with separate actuators~\cite{tsvetkov_novel_2022}, these approaches require deliberate leg movements to grasp and place objects. In contrast, our method achieves rapid scooping, tossing, and loading onto the back using only passive add-ons, maintaining high locomotion agility and minimizing mechanical complexity.

\section{Method}



In this section, we describe the overall Scoop-and-Toss framework, including the add-on design and the design and training of the expert policies and the meta-policy.

\subsection{Add-on Components}

\begin{figure}
  \centering
  \subfigure[]{
    \includegraphics[trim=0 50 310 300, clip, width=.40\linewidth]{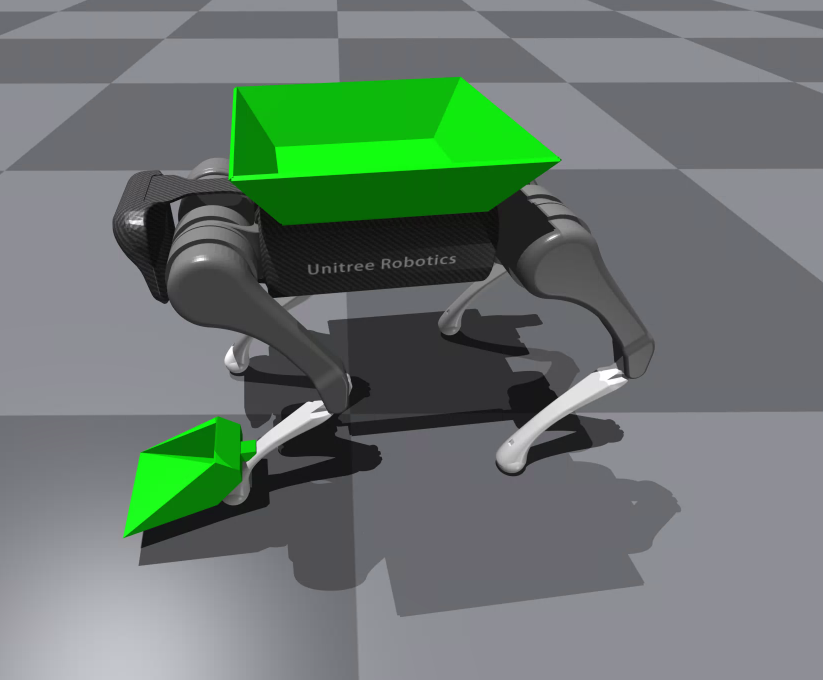}
    \label{fig:addon-scoop}
  }
  \subfigure[]{
    \includegraphics[trim=130 350 180 0, clip, width=.40\linewidth]{figures/addon-design.png}
    \label{fig:addon-storage}
  }
  \caption{Add-on components for Scoop-and-Toss: scoop (a) and collection tray (b).}
\end{figure}


We designed simple add-on components, consisting of a scoop and a tray, to enable the robot to pick up and load objects without additional actuators.

The scoop was custom-designed in Blender, inspired by various real-world shovel shapes (Fig.~\ref{fig:addon-scoop}).
Its dimensions are approximately 13.5\,cm in front width, 16.5\,cm in side width, and 7.5\,cm in height.
The weight was set to 200\,g, assuming a 3D-printed construction using a typical lightweight material.
The scoop is attached to one of the robot’s legs
and the angle between the scoop’s bottom surface and the calf link is approximately $135^\circ$.

The collection tray was implemented using the \textit{tray} model from Isaac Gym \cite{makoviychuk2021isaac}, scaled down to 25\% of its original size (Fig.~\ref{fig:addon-storage}).
Its dimensions are approximately 29\,cm in front and side width, and 7\,cm in height.
The weight was set to 400\,g.

\subsection{Expert Policies for Object Manipulation and Locomotion}

In the first stage, we decompose the overall task into two subtasks: picking up and tossing a single object using a scoop, and approaching a target object position.
We independently train a separate expert policy for each subtask to specialize in either object manipulation or navigation.

Both expert policies share the same input and output structure. Each policy receives the robot’s state, the previous action, the target object's position relative to the robot base, and the distance between the scoop and the object as input, and outputs target joint angles.
Further details of the input and output specifications can be found in Appendix~\ref{appd:expert-in-out}.

\paragraph{Scoop-and-Toss Policy ($\pi_\mathrm{scoop\_toss}$)}

In our design, the robot, without additional actuators, cannot firmly grasp objects with a gripper.  
Instead, it must pick them up using a simple scoop-shaped addon that merely supports the object.  
The robot must then toss the picked-up objects with appropriate speed and direction, making the task non-trivial.

Interestingly, these behaviors emerged from training a single policy with a unified reward function, without decomposing the task into separate scooping and tossing phases or designing distinct policies for each.

Specifically, the scoop-and-toss policy, $\pi_\mathrm{scoop\_toss}$, learns to pick up a single object and throw it into the collection tray, guided by the following reward formulation:
\begin{equation}
r_\mathrm{scoop\_toss} = r_\mathrm{toss} + b_\mathrm{load} + r_\mathrm{reg}.
\label{eq:r_st}
\end{equation}

The toss reward term $r_\mathrm{toss}$ is defined as:
\begin{equation}
r_\mathrm{toss} = w_1 \, h_\mathrm{obj} \, \exp(-w_2 \, d_\mathrm{obj}),
\end{equation}
where $h_\mathrm{obj}$ denotes the object’s height, and $d_\mathrm{obj}$ is its distance from the center of the tray floor.
This term encourages both lifting the object and directing it toward the tray center.
The exponential penalty on $d_\mathrm{obj}$ causes the reward to drop sharply as the object deviates from the target, while $h_\mathrm{obj}$ provides a positive contribution proportional to elevation.
Since $h_\mathrm{obj}$ increases meaningfully only after the object is supported by the scoop, the reward naturally induces a pick-up motion without requiring an explicit scooping bonus.
As a result, a smooth scoop-and-toss behavior emerges.
Furthermore, this multiplicative structure guides the policy to produce trajectories that are high enough for stable loading rather than simply maximizing height.
Curriculum learning was used to gradually increase the sampling radius for the object’s initial position around the scoop, allowing the robot to learn approaches before executing the scoop and toss.



In addition, $b_\mathrm{load}$ provides a bonus if the object is successfully loaded into the tray (i.e., if the object's center falls within the interior volume of the tray); otherwise, no bonus is given.
The additional regularization term $r_\mathrm{reg}$ is used to promote smooth and energy-efficient motion. 

An episode ends either when the object is successfully loaded and retained, when the object fails to be loaded within a curriculum-based time limit, or when the robot falls.
See Appendix~\ref{appd:training-pi-scoop-toss} and~\ref{appd:reward-terms} for training details and reward parameter settings.

\paragraph{Approach Policy ($\pi_\mathrm{approach}$)}

While $\pi_\mathrm{scoop\_toss}$ can scoop and load objects within a moderate range, it struggles with targets located behind or at sharp angles relative to the robot’s current orientation. These cases often require the robot to turn its body significantly before scooping, which introduces complex pose transitions that $\pi_\mathrm{scoop\_toss}$ was not trained to handle.
To address this, we introduce a second expert policy, $\pi_\mathrm{approach}$, dedicated to reliably moving the robot toward target object positions, particularly when significant reorientation is required.


The reward for $\pi_\mathrm{approach}$ is defined as follows:
\begin{equation}
r_\mathrm{approach} = w_3 \, \min {(\langle \mathbf{v}, \mathbf d_\mathrm{obj} \rangle, v_\mathrm{des})} + w_4 \, \exp( - w_5 \, \vert y_{obj} - y \vert ) + r_\mathrm{reg}.
\end{equation}
The first term encourages the robot to move toward the target object at a desired speed, following the velocity tracking term from \cite{cheng2023parkour}, where $\mathbf{v}$ is the robot’s linear velocity and $\mathbf d_\mathrm{obj}$ is the unit vector from the scoop to the object. The desired speed $v_\mathrm{des}$ is set to a constant value of 0.3\,m/s during training.
The second term encourages alignment between the robot’s heading $y$ and the object direction $y_\mathrm{obj}$, defined as the yaw angle of $\mathbf d_\mathrm{obj}$.
As in $r_\mathrm{scoop\_toss}$ (Equation~\ref{eq:r_st}), a regularization term is also applied.

At the beginning of each episode, the target object position is randomly sampled within a 5-meter radius from the robot base. The episode terminates once the scoop comes within 10\,cm of the object, indicating successful approach, or when the 20-second time limit is reached.
See Appendix~\ref{appd:reward-terms} for reward parameter settings.

\paragraph{Fine-Tuning for Smooth Policy Transitions}

To reduce abrupt behavior when switching between the two expert policies during meta-policy execution, we apply the Skill Transition-Based Initialization (STI) method proposed in \cite{PhysicsFC}. While STI was originally used during the sequential training of skills, we instead apply it in a post-training fine-tuning phase. Specifically, we collect intermediate states from one expert policy and use them as episode initial states when fine-tuning the other. This cross-initialization encourages the expert policies to generate behaviors that are more robust to transitions, improving the overall stability of meta-policy execution.
A detailed description of STI and how it is applied in our fine-tuning setup can be found in Appendix~\ref{appd:sti}.

\subsection{Meta-Policy for Multi-Object Collection}

After independently training the expert policies, we extend the environment to a multi-object setting, where the robot must repeatedly navigate and load multiple scattered objects.  
To achieve this, we train a meta-policy $\pi_\mathrm{meta}$ that selects between experts at each timestep.
The meta-policy shares the same input structure as the expert policies and outputs a probability distribution over them. At each timestep, the expert with the highest probability is selected for execution.



The reward for training the meta-policy $\pi_\mathrm{meta}$ is based on $r_\mathrm{scoop\_toss}$ (Equation~\ref{eq:r_st}), replacing the single-load bonus $b_\mathrm{load}$ with an accumulated bonus $b_\mathrm{load\_objs}$ proportional to the number of objects successfully loaded into the tray.
Formally, the reward is defined as:
\begin{equation}
r_\mathrm{meta} = r_\mathrm{toss} + b_\mathrm{load\_objs} + r_\mathrm{reg},
\end{equation}
where $b_\mathrm{load\_objs} = 100 \times N_\mathrm{loaded\_objs}$, and $N_\mathrm{loaded\_objs}$ denotes the cumulative number of objects loaded during an episode.
This formulation encourages strategies that maximize the total number of loaded objects.
Because both experts are conditioned on the object position, the robot generally moves in the correct direction regardless of which expert is selected. 
The role of the meta-policy is thus to learn when to switch between these experts—favoring $\pi_\mathrm{approach}$ when reaching the target requires significant reorientation or complex locomotion, and switching to $\pi_\mathrm{scoop\_toss}$ once the robot is well-positioned and can maximize the toss-related rewards.
The training environment includes the five cube-shaped objects randomly placed within a 5-meter radius, and selects the nearest uncollected object as the target at each timestep (see Appendix~\ref{appd:meta-env} for details).

\section{Experimental Results}
\label{sec:result}



\subsection{Training and Simulation Setup}

All experiments were conducted in the Isaac Gym simulator~\cite{makoviychuk2021isaac}.  
Both the expert policies and the meta-policy were trained using the Proximal Policy Optimization (PPO) algorithm~\cite{schulman_proximal_2017}.  
All policy networks consisted of three hidden layers with 256, 128, and 64 neurons, respectively, each employing the ELU activation function.


The scoop-and-toss policy was trained for 28k steps on a single RTX 3090 GPU in approximately 85 hours.
The approach policy, trained for 7k steps on a single RTX 4070 GPU, completed in under five hours.
The meta-policy was trained for 6k steps on the same RTX 4070 GPU after the expert policies were fixed, and the training took approximately 12 hours.
In addition, each expert policy was further fine-tuned using the STI (Appendix~\ref{appd:sti}).
Specifically, the scoop-and-toss policy was fine-tuned for 3k steps over five hours, and the approach policy for 5k steps over three hours.

Throughout training, a cube-shaped object with a side length of 4\,cm and a weight of 96\,g was used as the target object for all policies.

\subsection{Single-Object Scoop-and-Toss Performance}









% ...









\paragraph{Object Direction}
To assess the directional performance of $\pi_\mathrm{scoop\_toss}$, we divided the 360° space around the robot into eight 45° angular sectors:
0–45°, 45–90°, 90–135°, 135–180°, 180–225°, 225–270°, 270–315°, and 315–360°, with 0° defined as the robot’s forward-facing direction and positive angles measured counterclockwise (i.e., to the robot’s left).
For each sector, we conducted 100 trials in which the cube-shaped object’s initial position was randomly sampled within a 1.5\,m radius from the robot's scoop.

We then measured the success rates of the following stages in each angular sector, each defined by a specific criterion and considered a failure if not satisfied within 20 seconds:
\begin{itemize}
    \item \textbf{Approach Success:} The center of the scoop enters a 10\,cm radius around the object center.
    \item \textbf{Scoop Success:} A contact force is detected between the inner bottom surface of the scoop and the object.
    \item \textbf{Toss Success:} The object reaches a height exceeding that of the tray center.
    \item \textbf{Load Success:} The object’s center falls within the interior volume of the tray.
\end{itemize}

In addition to these success rates, we also report the following quantitative metric:
\begin{itemize}
    \item \textbf{Toss Height:} Maximum height achieved by the object during the toss.
\end{itemize}

\begin{table}[h]
\centering
\small
\caption{Scoop-and-toss performance metrics by angular sector of the object's initial position relative to the robot.
Angular Sector: range of angles between the robot’s forward-facing direction (0°) and the object’s initial position, measured counterclockwise (°). 
Approach/Scoop/Toss/Load: success rates for each stage (\%).  
Height (S/F): toss height for success/failure cases (m).  
Failure metrics (F) correspond to trials where the toss was successful but the object failed to land in the tray.
}
\label{tab:scoop_toss_angular}
\begin{tabular}{lccccccc}
\toprule
\textbf{Angular Sector(°)} & \textbf{Approach(\%)} & \textbf{Scoop(\%)} & \textbf{Toss(\%)} & \textbf{Load(\%)} & \textbf{Height(S/F)(m)} \\
\midrule
0--45 & 99 & 99 & 99 & 99 & 0.84 / - \\ 
45--90 & 99 & 99 & 99 & 99 & 0.75 / - \\ 
90--135 & 97 & 97 & 97 & 97 & 0.80 / - \\ 
135--180 & 55 & 52 & 52 & 52 & 0.76 / - \\ 
180--225 & 56 & 49 & 48 & 48 & 0.86 / - \\ 
225--270 & 99 & 98 & 98 & 98 & 0.85 / - \\ 
270--315 & 99 & 99 & 99 & 99 & 0.77 / - \\ 
315--360 & 99 & 97 & 97 & 97 & 0.77 / - \\ 
\bottomrule
\end{tabular}
\end{table}

As shown in Table~\ref{tab:scoop_toss_angular}, the policy achieved high success rates across all stages—Approach, Scoop, Toss, and Load—when the object was located within the 0–135° and 225–360° sectors, with success rates consistently above 97\%.
In contrast, performance dropped significantly in the 135–180° and 180–225° sectors, where the Approach success rate fell to around 55–56\%, and downstream stages showed similarly low success rates (48–52\%).
This indicates difficulty in handling objects located directly behind the scoop, likely due to limited turning capability and lack of explicit training on complex reorientation motions within $\pi_\mathrm{scoop\_toss}$.
Notably, once a toss was successful, the object was almost always loaded into the tray.
The consistently high performance across these angular sectors also implies that the policy is robust to variation in scoop-to-object distance within the tested range.
The \textit{Toss Height} remained relatively stable (0.75–0.86,m) across all angular sectors, with no significant difference even in the low-performing rear sectors.


\paragraph{Object Type}

\begin{figure}
  \centering
  \subfigure[]{
    \includegraphics[trim=0 200 0 150, clip, width=.60\linewidth]{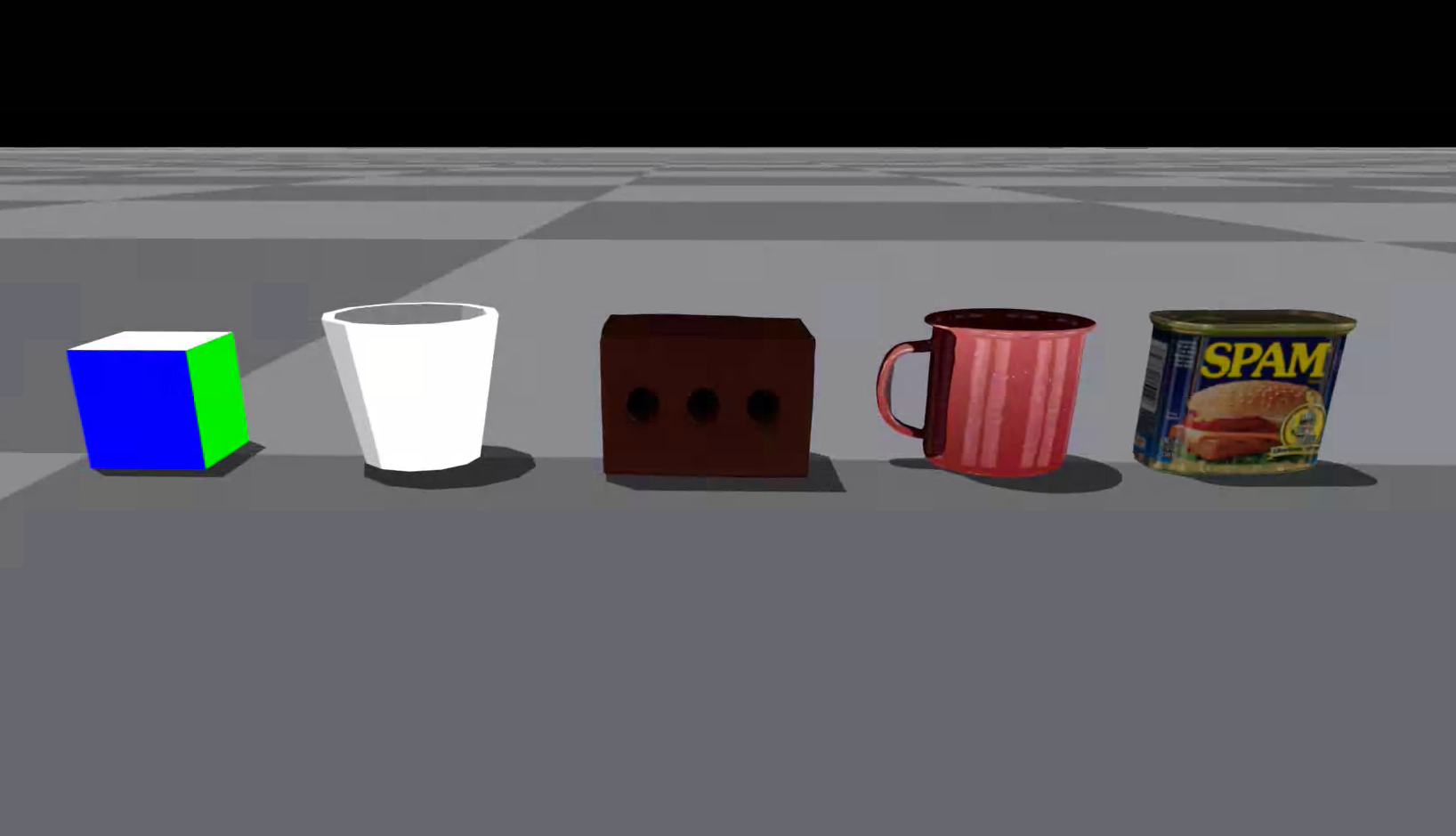}
    \label{fig:object-types}
  }
  \subfigure[]{
    \includegraphics[trim=500 215 400 450, clip, width=.36\linewidth]{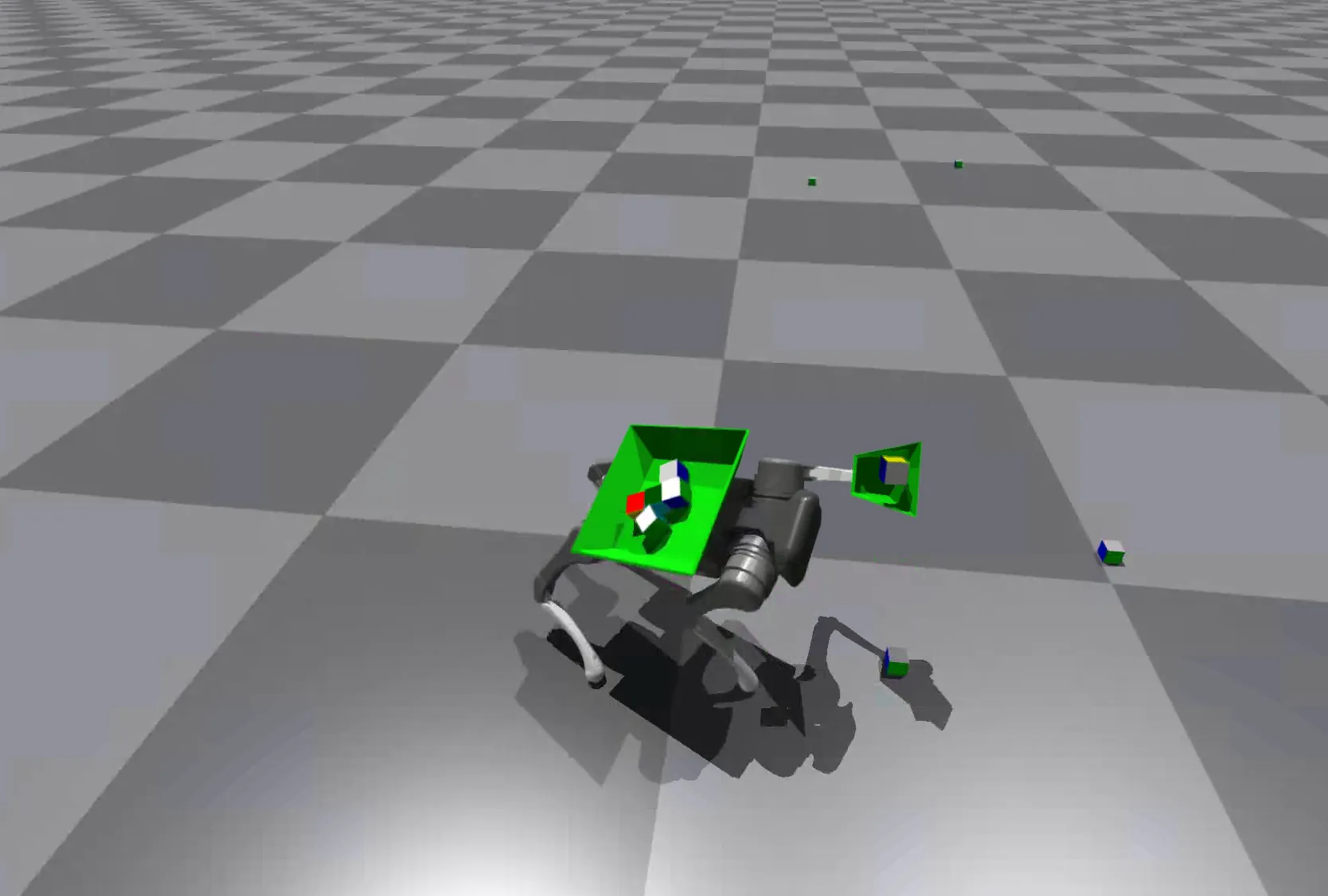}
    \label{fig:multi-objects}
  }
  \caption{(a) Objects used for the object type experiment: from left to right — Cube, Bucket, Foam Brick, Mug, and Potted Meat Can. (b) A multi-object collection trial using the meta-policy.}
\end{figure}

Although $\pi_\mathrm{scoop\_toss}$ was trained solely on the cube object, it successfully generalized to various unseen object types at test time.
The test objects, sourced from IsaacGym examples, were rescaled to fit within the scoop and had their masses adjusted accordingly (Figure~\ref{fig:object-types}).
For this evaluation, we conducted an ablation study under the same conditions as Table~\ref{tab:scoop_toss_angular}, restricting object angles to the ±135° frontal range, where baseline performance remained stable.

\begin{table}[h]
\centering
\small
\caption{Scoop-and-toss performance across different object types.  
Metrics are defined as in Table~\ref{tab:scoop_toss_angular}.}
\label{tab:scoop_toss_object_types}
\begin{tabular}{lccccc}
\toprule
\textbf{Object Type (Mass)} & \textbf{Approach (\%)} & \textbf{Scoop (\%)} & \textbf{Toss (\%)} & \textbf{Load (\%)} & \textbf{Toss Height (S/F) (m)} \\
\midrule
Cube (96\,g) & 98 & 98 & 98 & 98 & 0.76 / -- \\
Bucket (80\,g) & 97 & 95 & 92 & 87 & 0.85 / 0.85 \\
Mug (80\,g) & 96 & 62 & 48 & 11 & 0.82 / 0.81 \\
Foam Brick (30\,g) & 98 & 87 & 89 & 79 & 0.90 / 0.92 \\
Potted Meat Can (220\,g) & 97 & 93 & 93 & 93 & 0.79 / -- \\
\bottomrule
\end{tabular}
\end{table}

As shown in  Table~\ref{tab:scoop_toss_object_types}, The potted meat can, despite being the heaviest object (220\,g), achieved over 90\% success at all stages.
Its cubic shape likely allowed the policy—trained on similarly shaped objects—to generalize well despite the increased mass.
The foam brick also shares a box-like shape but is extremely light.
While it was easy to scoop and toss, its low mass often caused it to travel farther than intended, occasionally landing near or beyond the rear edge of the tray and resulting in load failures.
The bucket, with a rounded body, showed slightly lower success than the cube or meat can, suggesting some degradation in performance for non-cubic shapes.
Interestingly, the mug—despite having a similar rounded body and mass to the bucket—performed much worse, with only 11\% load success.
This is likely due to its handle, which interfered with rolling and alignment within the scoop, making pickup and toss more unstable.

\paragraph{Ablation by Reward Design}


To validate the reward design for $\pi_\mathrm{scoop\_toss}$, we performed an ablation study using the same setup as in the object type experiment.




\begin{table}[h]
\centering
\small
\caption{Performance of $\pi_\mathrm{scoop\_toss}$ under different reward ablation settings.}
\label{tab:scoop_toss_ablation}
\begin{tabular}{lccccccc}
\toprule
\textbf{Ablation Setting} & \textbf{Approach(\%)} & \textbf{Scoop(\%)} & \textbf{Toss(\%)} & \textbf{Load(\%)} & \textbf{Height(S/F)(m)} \\
\midrule

Baseline $\pi_\mathrm{scoop\_toss}$ & 98 & 98 & 98 & 98 & 0.76 / - \\ 
No Height Reward              & 20  & 0  & 0  & 0 & - / - \\ 
No ExpDist Reward             & 20  & 0  & 1  & 0 & - / 0.89 \\ 
No Load Bonus                 & 27 & 15 & 14 & 11  & 1.07 / 1.00 \\ 

\bottomrule
\end{tabular}
\end{table}

As shown in Table~\ref{tab:scoop_toss_ablation}, the baseline policy achieved near-perfect performance across all stages.  
In all ablation conditions, the drop in load success hindered curriculum progression—structured around load outcomes—and prevented the policy from learning reliable approach behavior, resulting in low Approach success rates.  
Removing either the height reward or the exponential distance reward led to complete failure, underscoring their importance in shaping overall behavior and enabling successful tosses.  
In the \textit{No ExpDist} case, a single toss occurred where the robot failed to scoop the object and instead struck it with the scoop edge, flinging it away—a result of optimizing for height without regard for proximity.  
Removing the load bonus degraded performance across all stages.  
Lacking incentive for accurate landings, the policy tended to overemphasize toss height at the expense of stable scooping and control.

\subsection{Multi-Object Meta-Policy Performance}

We evaluated $\pi_\mathrm{meta}$ in a multi-object environment with the ten cube-shaped objects randomly placed within a 5-meter radius.  
The robot repeatedly scooped and loaded the objects into the tray by dynamically switching between $\pi_\mathrm{scoop\_toss}$ and $\pi_\mathrm{approach}$ (Figure~\ref{fig:multi-objects}).
Each episode terminated either after 100 seconds or when all objects had been successfully loaded into the tray.  
We ran 100 episodes in total and measured the average number of objects loaded per episode and the average time per object.


\begin{table}[h]
\centering
\small
\caption{Performance comparison between the baseline $\pi_\mathrm{meta}$ and ablated variants in the multi-object task.  
Loaded Objects: average number of objects loaded per episode.  
Time per Object: average time taken to load one object (s).}
\label{tab:meta_ablation}
\begin{tabular}{lcc}
\toprule
\textbf{Model} & \textbf{Loaded Objects} & \textbf{Time per Object (s)} \\
\midrule
Baseline $\pi_\mathrm{meta}$ & 7.22 & 13.1 \\
No STI & 0 & 0 \\
No Extra Bonus & 0 & 0 \\
$\pi_\mathrm{scoop\_toss}$ Only & 2.17 & 4.45 \\
\bottomrule
\end{tabular}
\end{table}

As shown in Table~\ref{tab:meta_ablation}, the baseline $\pi_\mathrm{meta}$ loaded 7.22 objects per episode on average, with 13.1 seconds per object.
In contrast, both the \textit{No STI} and \textit{No Extra Bonus} variants failed to load any objects. In both cases, the policy effectively collapsed to using only $\pi_\mathrm{approach}$. Without STI, the policy tended to avoid switching altogether, likely due to instability from abrupt transitions. In the \textit{No Extra Bonus} condition, the robot did not attempt any scoop-and-toss actions, suggesting that the absence of multi-object incentives discouraged engagement with the scooping task.
In the \textit{$\pi_\mathrm{scoop\_toss}$-Only} condition, the robot loaded only 2.17 objects per episode. As shown in Table~\ref{tab:scoop_toss_angular}, the policy struggles with objects located behind the robot, and this inability to handle difficult orientations contributes significantly to the drop in overall performance.

\subsection{Interactive Joystick Control}

We additionally verified that our system supports interactive control by conducting a joystick-based experiment.  
In this setting, the user steered the robot by specifying a desired movement direction using a joystick. 
A target position was then generated by projecting a fixed distance along the input direction and used as input to $\pi_\mathrm{approach}$.  
The user could also trigger the scoop-and-toss action using a button input.  
This experiment demonstrates that the learned expert policies can be naturally integrated into simple interactive control schemes.  
Please refer to the supplementary video for demonstration details.



\section{Conclusion and Discussion}

In this work, we introduced a simple yet effective Scoop-and-Toss framework for quadruped robots, consisting of a scoop and a collection tray add-on.  
Our system trains the scoop-and-toss policy for picking up and tossing objects, the approach policy for goal-directed locomotion, and the meta-policy that dynamically switches between these experts to collect multiple scattered objects.

Extensive evaluations demonstrated the effectiveness of the proposed framework.  
The scoop-and-toss policy learned to coordinate locomotion and tossing through a unified reward, showing robust performance in non-rear directions, generalization to unseen object types.
The meta-policy effectively combined locomotion and scooping skills, enabling the robot to collect multiple objects by automatically selecting the nearest uncollected object as the target.

These results highlight the broader potential of quadruped robots to perform dynamic loco-manipulation tasks by leveraging the inherent agility and strength of their legs.  
Our findings suggest that minimal mechanical modifications, combined with hierarchical policy learning, can unlock new capabilities in quadruped systems, advancing their role from pure locomotion toward more complex, dynamic interaction with the environment.

\section{Limitations}

Despite the demonstrated capabilities, our framework has several limitations.

First, all evaluations were conducted in simulation, and we have not yet validated the framework on a physical robot.  
Although the add-on components were designed with real-world feasibility in mind, deploying the policies on hardware may require additional considerations for perception noise, actuation delay, and domain transfer.

Second, although the policy was trained only on a single object type, we evaluated it on previously unseen objects with varied shapes, sizes, and masses, and observed promising generalization.  
However, the policy does not explicitly infer or utilize object properties such as shape or mass.  
Incorporating mechanisms to estimate such properties from observation could improve robustness and enable more adaptive behavior across object categories.

In addition, our framework assumes access to accurate object positions.  
In real-world settings, sensor noise and partial observations could impact performance, requiring more robust perception and control strategies.

Furthermore, the approach policy $\pi_\mathrm{approach}$ was trained with a fixed target speed.  
Allowing the policy to receive speed as an explicit input could enable more flexible and adaptive navigation, especially in scenarios requiring dynamic modulation of approach velocity.

Lastly, our experiments assumed static objects and flat terrain conditions.  
Extending the framework to handle dynamic object motion and uneven surfaces would pose additional challenges for balance, foot placement, and real-time adaptation, but also open up opportunities for further generalization.

Addressing these limitations will be crucial for advancing the Scoop-and-Toss framework toward practical deployment in real-world environments.



\clearpage
\acknowledgments{If a paper is accepted, the final camera-ready version will (and probably should) include acknowledgments. All acknowledgments go at the end of the paper, including thanks to reviewers who gave useful comments, to colleagues who contributed to the ideas, and to funding agencies and corporate sponsors that provided financial support.}



\bibliography{ref}  

\appendix

\section{Expert Policy Input and Output Specifications}
\label{appd:expert-in-out}

Both expert policies, $\pi_\mathrm{scoop\_toss}$ and $\pi_\mathrm{approach}$, share a common input structure, consisting of:
\begin{itemize}
    \item Robot state $\mathbf{s}_t \in \mathbb{R}^{33}$: projected gravity, base angular velocity, base acceleration, joint positions, and joint velocities.
    \item Previous action $\mathbf{a}_{t-1} \in \mathbb{R}^{12}$.
    \item Object position $\mathbf{p}_t^\mathrm{obj} \in \mathbb{R}^3$ relative to the robot base.
    \item  Distance between the scoop and the object $d_t^\mathrm{obj} \in \mathbb{R}^1$.
\end{itemize}

The output action $\mathbf{a}_t \in \mathbb{R}^{12}$, shared by both policies, specifies target joint angles relative to the default joint configuration, representing the desired displacement from the default posture.

\section{Training Details for Scoop-and-Toss Policy}
\label{appd:training-pi-scoop-toss}

\subsection{Curriculum Learning}

To facilitate the emergence of the scooping behavior, the policy $\pi_\mathrm{scoop\_toss}$ was initially trained with the object placed approximately 5\,cm in front of the scoop tip.  
Curriculum learning was applied by gradually expanding the sampling radius around this initial placement point as training progressed.
At each curriculum level, the object’s position was sampled within a circular area whose radius increased by 5\,cm once the policy achieved at least 10 episodes with a success rate exceeding 80\%.
Through this process, training continued until the sampling radius reached 1.5\,m around the initial placement point, allowing $\pi_\mathrm{scoop\_toss}$ to learn to scoop objects placed at increasingly distant positions.

\subsection{Episode Termination Conditions}

Episodes terminate under any of the following conditions:

\begin{itemize}
    \item The object is successfully loaded into the tray and retained for 5\,s.

    \item The object fails to be loaded within a time limit that increases progressively as training advances.
    Specifically, the time limit starts at 1\,s and is incrementally extended by 0.5\,s at each curriculum update, encouraging the robot to first master quick scooping behaviors and progressively handle more deliberate and stable actions.

    \item The robot falls during the episode.

    \item A 20-second timeout is reached, which accounts for cases where the object is initially loaded into the tray but dropped before satisfying the 5-second retention requirement (without the robot falling).
\end{itemize}

\section{Reward Terms and Parameters}
\label{appd:reward-terms}

\subsection{Bonus Values and Weight Parameters}

The specific values assigned to the reward terms were determined experimentally.
For $\pi_\mathrm{scoop\_toss}$, we used $b_\mathrm{load} = 100.0$, $w_1 = 30.0$, and $w_2 = 4.0$.  
For $\pi_\mathrm{approach}$, we used $w_3 = 3.5$, $w_4 = 1.0$, and $w_5 = 1.0$.

\subsection{Regularization Term}

Both policies, $\pi_\mathrm{scoop\_toss}$ and $\pi_\mathrm{approach}$, incorporate an additional regularization term $r_\mathrm{reg}$ to promote smooth and energy-efficient motion, defined as:
\begin{equation}
r_\mathrm{reg} = w_6 \| \ddot{\mathbf \varphi} \|^2 + w_7 \| \mathbf a_{t-1} - \mathbf a_t \|^2 + w_8 \| \mathbf \tau \|^2,
\end{equation}
where $\ddot{\mathbf \varphi}$ is the joint acceleration, $\mathbf a$ is the output action, and $\mathbf \tau$ is the joint torque for all joints.

The weights for the regularization components were determined experimentally as follows:
for $\pi_\mathrm{scoop\_toss}$, $w_6 = -1 \times 10^{-3}$, $w_7 = -1 \times 10^{-4}$, and $w_8 = -3 \times 10^{-5}$;
for $\pi_\mathrm{approach}$, $w_6 = -1 \times 10^{-2}$, $w_7 = -3 \times 10^{-3}$, and $w_8 = -1 \times 10^{-5}$.

\section{Fine-Tuning with Skill Transition-Based Initialization (STI)}
\label{appd:sti}

Skill Transition-Based Initialization (STI)~\cite{PhysicsFC} is a training strategy for improving the robustness and fluidity of transitions between skill policies in multi-skill reinforcement learning.
Instead of initializing episodes from a fixed distribution or random reset, STI samples initial states from intermediate steps in other skill executions. This exposes the policy to state distributions it is likely to encounter after a skill switch.
Intermediate states are collected from rollouts of trained policies and stored in skill-specific buffers. States are saved at points where transitions are likely, such as after key events or at random steps, depending on the task.

We adopt STI as a post-training fine-tuning step to improve the transition stability between two independently trained expert policies, $\pi_\mathrm{approach}$ and $\pi_\mathrm{scoop\_toss}$.
The STI buffer for each policy is constructed by collecting 10,000 robot states sampled at random time points during rollouts of the respective trained policy.
Goal inputs are sampled as in initial training, and only the robot’s state  is stored.
During fine-tuning, each episode is initialized using a randomly selected state from the other expert’s STI buffer.

\section{Meta-Policy Environment Setup}
\label{appd:meta-env}

The environment is initialized with the five cube-shaped objects, randomly placed within a 5-meter radius centered at the robot base.
Unlike the expert policies for single-object tasks, $\pi_\mathrm{meta}$ selects one object as the target at a time. The target is defined as the nearest uncollected object relative to the robot, and a new target is selected whenever the current one is successfully loaded.
The position and distance of the current target object are provided as inputs to $\pi_\mathrm{meta}$, which selects either $\pi_\mathrm{approach}$ or $\pi_\mathrm{scoop\_toss}$ for execution. The selected expert policy receives the same input information to generate the corresponding action.

\end{document}